\documentclass[12pt,reqno]{amsart}
\usepackage{amsmath}
\usepackage{amsthm}
\usepackage{amssymb}
\usepackage{algorithm2e}
\usepackage[all]{xy}
\usepackage{color}
\usepackage{graphicx}
\usepackage{float}
\usepackage{hyperref}
\usepackage{orcidlink}
\usepackage{tikz}
\usepackage{forest}
\usepackage{xcolor}
\usepackage{subcaption}
\usetikzlibrary{trees}
\usepackage{lineno}
\usetikzlibrary{positioning,arrows.meta}
\usetikzlibrary{calc}
\usepackage{booktabs}
\usepackage{capt-of}

\RestyleAlgo{ruled}

\definecolor{blue_2007}{RGB}{175, 209, 230}
\definecolor{blue_1687}{RGB}{191, 216, 236}
\definecolor{blue_2063}{RGB}{172, 208, 230}
\definecolor{blue_1766}{RGB}{187, 214, 235}
\definecolor{blue_3441}{RGB}{89, 162, 207}
\definecolor{blue_754}{RGB}{222, 235, 247}
\definecolor{blue_2038}{RGB}{173, 208, 230}
\definecolor{blue_690}{RGB}{225, 236, 247}
\definecolor{blue_3858}{RGB}{67, 147, 198}
\definecolor{blue_4385}{RGB}{44, 124, 187}
\definecolor{blue_2056}{RGB}{172, 208, 230}
\definecolor{blue_4725}{RGB}{30, 110, 178}
\definecolor{blue_1726}{RGB}{190, 215, 236}
\definecolor{blue_2456}{RGB}{149, 197, 223}
\definecolor{blue_651}{RGB}{226, 237, 248}

\theoremstyle{definition}

\theoremstyle{remark}

\numberwithin{equation}{section}
\numberwithin{theorem}{section}

\graphicspath{ {images/} }

\begin{document}

	\title[Soft Computing Approaches for Predicting Shade-Seeking Behaviour in Dairy Cattle under Heat Stress: A Comparative Study of Random Forests and Neural Networks]{Soft Computing Approaches for Predicting Shade-Seeking Behaviour in Dairy Cattle under Heat Stress: A Comparative Study of Random Forests and Neural Networks}

	\author[Sanjuan, S.]{S. Sanjuan\orcidlink{0009-0001-5310-2559}}
	\address{%
		Instituto Universitario de Matem\'atica Pura y Aplicada\\
		Universitat Polit\`ecnica de Val\`encia\\
		46022 Valencia\\
		Spain}
	\email{\textcolor[rgb]{0.00,0.00,0.84}{ssansil@upvnet.upv.es} (S.S.)}

	\author[Méndez, D.A.]{Daniel Alexander Méndez Reyes\orcidlink{0000-0001-6414-305X}}
	\address{Instituto Ciencia y Tecnología Animal\\
		Universitat Polit\`ecnica de Val\`encia\\
		46022 Valencia\\
		Spain}
	\email{\textcolor[rgb]{0.00,0.00,0.84}{damenre@upvnet.upv.es} (D.A.M.)}

	\author[Arnau, R.]{R. Arnau\orcidlink{0000-0003-2544-8875}}
	\address{%
		Instituto Universitario de Matem\'atica Pura y Aplicada\\
		Universitat Polit\`ecnica de Val\`encia\\
		46022 Valencia\\
		Spain}
	\email{\textcolor[rgb]{0.00,0.00,0.84}{ararnnot@posgrado.upv.es} (R.A.)}
	
	\author[Calabuig, J.M.]{J. M. Calabuig\orcidlink{0000-0001-8398-8664}}
	\address{Instituto Universitario de Matem\'atica Pura y Aplicada\\
		Universitat Polit\`ecnica de Val\`encia\\
		46022 Valencia\\
		Spain}
	\email{\textcolor[rgb]{0.00,0.00,0.84}{jmcalabu@mat.upv.es} (J.M.C.)}
	
	\author[Diaz, X.]{Xabier Diaz de Otalora Aguirre\orcidlink{0000-0003-4841-079X}}
	\address{Instituto Ciencia y Tecnología Animal\\
		Universitat Polit\`ecnica de Val\`encia\\
		46022 Valencia\\
		Spain}
	\email{\textcolor[rgb]{0.00,0.00,0.84}{xdiadeo@upv.es} (X.D.)}

	\author[Estellés, F.]{Fernando Estellés Barber\orcidlink{0000-0001-6774-6075}}
	\address{Instituto Ciencia y Tecnología Animal\\
		Universitat Polit\`ecnica de Val\`encia\\
		46022 Valencia\\
		Spain}
	\email{\textcolor[rgb]{0.00,0.00,0.84}{feresbar@upv.es} (F.E.)}

	
	
	\date{June 22, 2025}

	\maketitle

	
	\begin{abstract}
		Heat stress is one of the main welfare and productivity problems faced by dairy cattle in Mediterranean climates. In this study, we approach the prediction of the daily shade-seeking count as a non-linear multivariate regression problem and evaluate two  soft computing algorithms---Random Forests and Neural Networks---trained on high-resolution behavioral and micro-climatic data collected in a commercial farm in Titaguas (Valencia, Spain) during the $2023$ summer season.
		The raw dataset ($6907$ daytime observations, $5$-$10$ min resolution) includes the number of cows in the shade, ambient temperature and relative humidity. From these we derive three  features: current Temperature--Humidity Index (THI), accumulated daytime THI, and mean night-time THI. To evaluate the models’ performance a $5$-fold cross-validation is also used. Results show that both soft computing models outperform a single Decision Tree baseline. The best Neural Network ($3$ hidden layers, $16$ neurons each, learning rate  $=10^{-3}$) reaches an average RMSE of $14.78$, while a  Random Forest ($10$ trees, depth  $= 5$) achieves $14.97$ and offers best interpretability. Daily error distributions reveal a median RMSE of $13.84$ and confirm that predictions deviate less than one hour from observed shade-seeking peaks.
		These results demonstrate the suitability of soft computing, data-driven approaches  embedded in an applied-mathematical feature framework for modeling noisy biological phenomena, demonstrating their value as low-cost, real-time decision-support tools for precision livestock farming under heat-stress conditions.
	\end{abstract}
	


	\section{Introduction}\label{sec:intro}
	
	In the context of livestock productivity, mathematical modeling allows understanding and quantifying the complex interaction of environmental variables and physiological responses. Taking advantage of well-established {\bf principles of mathematical analysis and modeling}, this work aims to establish a robust mathematical framework that captures the dynamics of temperature-humidity indices  and their relationship to animal welfare. This framework not only facilitates a deeper theoretical understanding, but also provides the necessary structure to integrate machine learning methodologies. Through this combination, we aim to improve predictive capacity and practical knowledge to improve livestock management in different climatic conditions.
	
	Artificial Intelligence (AI) is a field of Computer Science that focuses on creating systems capable of performing tasks that normally require human intelligence. These systems include a wide range of tasks ranging from learning, perception, or reasoning to problem solving, image recognition, or decision making. Moreover, {\bf Machine learning} (ML) techniques---supervised, unsupervised and reinforcement---are now standard tools across science and engineering (such as, for instance, in the Economy \cite{CALABUIG2020172} or Sport Sciences \cite{notari2023goal}, but also in the case of the livestock sector \cite{ani14101497, SLOB2021105237}).
	In this sense, {\bf soft computing} is the collective term coined by Zadeh (1994) for computational paradigms---fuzzy logic, neural networks, evolutionary algorithms and their hybrids---that trade exactness for robustness and tolerance to uncertainty. Unlike "hard computing" methods based on exact logic, soft computing approaches are designed to handle noisy, incomplete or vaguely defined data while delivering solutions that are "good enough" for complex real-world problems (\cite{Zadeh}).
	Biological processes rarely unfold in the orderly, deterministic manner that traditional computational methods assume, but are conditioned by stochastic fluctuations in the environment, sensor noise, and the intrinsic heterogeneity of living organisms. The behavior of dairy cattle is a clear example: even under identical thermal loads individual cows respond differently, and camera-based counts of shade-seeking animals are unavoidably imprecise. Because soft computing was explicitly conceived to "compute with words, perceptions and uncertain data", it offers a natural fit for this problem domain. Techniques such as Random Forests and Neural Networks tolerate incomplete or noisy inputs, capture nonlinear interactions without requiring a fully specified physiological model, and produce approximate---but operationally useful---outputs that can drive real-time management decisions. By leveraging this tolerance to imprecision, soft computing models provide robust, low-cost tools for precision livestock farming, where the objective is not perfect prediction of every individual but reliable, adaptive guidance under biologically variable conditions.

	While ML techniques are widely applied in {\bf livestock farming}, most studies focus on milk production  (\cite{JI2022186}), disease prediction (\cite{LUO2023106059}), or feed optimization  (\cite{ani14101497}). However, few efforts have targeted behavioral responses to environmental stress, such as shade-seeking, despite its critical importance for mitigating heat stress in animals. This paper addresses this gap by developing predictive models adapted for livestock management in Mediterranean regions, which are particularly vulnerable to heat stress. In our case, we will use three well-known algorithms of supervised learning: Random Forest (as a generalization of a Decision Tree, which will be the first) and Neural Networks. Random Forest has a range of applications in livestock farming, such as, for instance, prediction of milk production (\cite{JI2022186}), disease detection (\cite{LUO2023106059}), meat quality classification (\cite{SHIRZADIFAR2023100317}), feed optimization (\cite{ani14101497}), fertility and reproduction prediction (\cite{HEMPSTALK20155262}) or environmental health management (\cite{VALLETTA2017203}). Neural Networks have also been widely used in similar contexts (see, for instance, \cite{GRZESIAK200669,NAQVI2022106618,LIN2023107638}). Details  of these three soft computing algorithms 
	can be found in Section \ref{sec:meth}.

	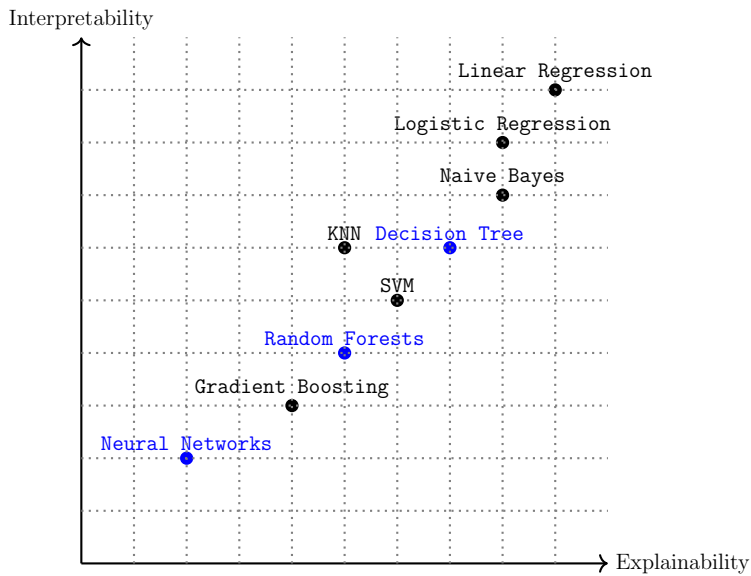
\begin{figure}[h]
		\centering \begin{tikzpicture}[thick,scale=0.75, every node/.style={scale=0.8}]
			\draw[->] (0,0) -- (10,0) node[right] {Explainability};
			\draw[->] (0,0) -- (0,10) node[above] {Interpretability};
			\filldraw[black] (9,9) circle (3pt) node[anchor=south] {\tt Linear Regression};
			\filldraw[blue] (7,6) circle (3pt) node[anchor=south] {\tt Decision Tree};
			\filldraw[black] (8,8) circle (3pt) node[anchor=south] {\tt Logistic Regression};
			\filldraw[blue] (5,4) circle (3pt) node[anchor=south] {\tt Random Forests};
			\filldraw[blue] (2,2) circle (3pt) node[anchor=south] {\tt Neural Networks};
			\filldraw[black] (4,3) circle (3pt) node[anchor=south] {\tt Gradient Boosting};
			\filldraw[black] (6,5) circle (3pt) node[anchor=south] {\tt SVM};
			\filldraw[black] (5,6) circle (3pt) node[anchor=south] {\tt KNN};
			\filldraw[black] (8,7) circle (3pt) node[anchor=south] {\tt Naive Bayes};
			\draw[dotted, color=gray] (0,1) -- (10,1);
			\draw[dotted, color=gray] (0,2) -- (10,2);
			\draw[dotted, color=gray] (0,3) -- (10,3);
			\draw[dotted, color=gray] (0,4) -- (10,4);
			\draw[dotted, color=gray] (0,5) -- (10,5);
			\draw[dotted, color=gray] (0,6) -- (10,6);
			\draw[dotted, color=gray] (0,7) -- (10,7);
			\draw[dotted, color=gray] (0,8) -- (10,8);
			\draw[dotted, color=gray] (0,9) -- (10,9);
			\draw[dotted, color=gray] (1,0) -- (1,10);
			\draw[dotted, color=gray] (2,0) -- (2,10);
			\draw[dotted, color=gray] (3,0) -- (3,10);
			\draw[dotted, color=gray] (4,0) -- (4,10);
			\draw[dotted, color=gray] (5,0) -- (5,10);
			\draw[dotted, color=gray] (6,0) -- (6,10);
			\draw[dotted, color=gray] (7,0) -- (7,10);
			\draw[dotted, color=gray] (8,0) -- (8,10);
			\draw[dotted, color=gray] (9,0) -- (9,10);
			
		\end{tikzpicture}
		\caption{Relationship between interpretability and explainability of several supervised ML algorithms. The algorithms highlighted in blue are the ones utilized in this study.}
		\label{fig:exp_int}
	\end{figure}
	
	In this context, a relevant difference between these two types of algorithms (Random Forest and Neural Networks) lies in the concepts of {\bf explainability} and {\bf interpretability}. 
	Explainability and interpretability are key concepts when it comes to understanding and trusting models. On the one hand explainability refers to the ability of the model to provide understandable explanations of its behavior and results. In other words, explainability focuses on providing details and reasons about how and why a prediction was produced.
	On the other hand, interpretability refers to the ease with which a human being can understand the reasons behind a model's predictions. 
	Figure \ref{fig:exp_int} positions the three algorithms studied---Decision Tree, Random Forest and Neural Network---along the interpretability-explainability spectrum. Information about interpretability and explainability can be found in \cite{molnar2022}.

	We end this first section by talking about the main feature on which our dataset (that will be explained at the beginning of the second section): the {\bf Temperature-Humidity Index} (THI). 
	Mitigating climate change impacts on dairy cattle production systems is essential for the sector's sustainability, especially as rising temperatures increase the risk of heat stress and negatively impact animal welfare and productivity, particularly in Mediterranean regions. In response, AI integrated with precision livestock farming (PLF) technologies enables detailed analysis of individual data, supporting the adoption and assessment of targeted strategies to reduce heat stress. 
	The THI index, refers to a measure used to evaluate the combined effects of temperature and humidity on the well-being and performance of animals, particularly livestock such as dairy cattle, poultry, and pigs. It is commonly used in agriculture and animal husbandry to assess the risk of heat stress, which can significantly impact animal health, productivity, and welfare (\cite{polsky2017invited}). To the best of our knowledge, this is the first time that this type of models has been used to predict thermal stress based on behavioral changes in animals and THI data, conducting non-invasive methods for data capture using computer vision.
	
	In summary, in this paper we therefore aim to select the most appropriate AI approach to predict the number of cows under the shadow under different THI conditions. To this end, Decision Trees, Random Forest and Neural Networks will be assessed. Animals exposed to high temperatures often exhibit behavioral adaptations to alleviate thermal discomfort. Among these, seeking shade is a primary response, even preferred over other cooling strategies like sprinklers or showers. Of course, providing shaded areas can effectively reduce the thermal load on animals, enhancing their comfort and performance. Moreover, a study by Schütz et al. (2010) highlighted that dairy cattle prioritize access to shade under heat stress conditions, underscoring its importance as a mitigation strategy (\cite{Schutz2011-kn}). 
	
	Our methodology incorporates novel features derived from the raw data, such as accumulated THI and the previous night’s average THI, to capture both immediate and cumulative effects of heat stress on cow behavior. Additionally, this study is among the first to compare Random Forests and Neural Networks in this context, offering insights into the trade-offs between accuracy and interpretability for real-world farm management. The use of $5$-fold cross-validation ensures robust model evaluation, minimizing bias and variance across a limited dataset.
	
	The remainder of the paper describes the dataset and models (Section \ref{sec:meth}), presents the results (Section \ref{sec:results}), 
	discusses their implications  (Section \ref{sec:discussion}) and summarises the main conclusions  (Section \ref{sec:conclusions}).

	\section{Materials and Methods}\label{sec:meth}
	
	\subsection{Dataset description and processing}

	The dataset for our study originates from a farm located in Titaguas, Val\`encia in Spain. This farm includes a feedlot with a shaded area in the center, monitored by three cameras that count the number of cows within the shaded area. The dataset consists of observations taken every 5-10 minutes during the summer of 2023, spanning from July 11th, 2023 to October 16th, 2023. The observations differ between day and night.
	
	During the day (from 07:00 to 21:00), each observation includes the number of cows in the shade, the temperature (in degrees Celsius), the relative humidity (in percent) and the exact time of observation.
	As for the nighttime (from 21:00 to 07:00), each observation includes the temperature, relative humidity and the exact time of observation, since the number of cows in the shade is meaningless. Similar to the daytime data, temperature and relative humidity are recorded to track nighttime environmental conditions.
	
	From these data, we have derived new variables for use in the models, in addition to the number of cows in the shade and the time. Firstly, the time variable has been transformed into a continuous real value for model training. This transformation allows the model to process time as a numerical feature rather than a categorical one. Secondly, with temperature and relative humidity, we calculate the Temperature-Humidity Index (THI). There are different formulas to calculate THI depending on the context, the type of animal and the region (see \cite{Habeeb2018TemperatureHumidityIA} and the references therein). In our case, the formula we are using, originally proposed by the National Research Council (NRC) in 1971 (\cite{NRC1971}), is:
	\begin{equation}\label{THI}
		\text{THI} = (1.8 \times \text{T}_{\text{db}} + 32) - \big((0.55 - 0.0055 \times \text{RH}) \times (1.8 \times \text{T}_{\text{db}} - 26)\big),
	\end{equation}  
	where $\text{T}_{\text{db}}$ is dry bulb temperature in Celsius and $\text{RH}$ is the relative humidity in decimal form. This index serves as a key indicator in our models, quantifying the combined effect of temperature and humidity on the cows' comfort. Furthermore, we derived two additional variables from the THI: the previous night's THI, calculated as the average of the previous night, and the accumulated THI, which is the average from the first hour of the day (07:00) up to the current observation time.
	
	The dataset includes the following variables (columns):
	\begin{itemize}
		\item Number of cows in the shade.
		\item Exact time of observation.
		\item Current THI.
		\item Average THI of the previous night.
		\item Accumulated THI.
	\end{itemize}
	
	Initially the data spans from July 11th, 2023 to October 16th, 2023, covering a total of 98 days, we were ultimately left with 75 days of data due to some days not being correctly recorded.
	This results in a total of $6907$ observations.
	Each day has distinct characteristics or variables, so we employed a \textbf{cross-validation} with $5$ folds for model validation. This choice is motivated by several reasons, including the need to assess the model's performance reliably with a limited dataset and to ensure that the model generalizes well to unseen data  (see \cite{kohavi1995study, stone1974cross}).
	
\begin{figure}[h]
		\resizebox{\textwidth}{!}{
			\begin{tikzpicture}[node distance=0mm,minimum height=1cm,outer sep=2mm,scale=0.9,>=Latex,font=\tiny, 
				indication/.style={minimum height=0cm,outer sep=0mm},
				oneblock/.style={transform shape,minimum width=1cm,draw},
				fullset/.style={transform shape,minimum width=5cm,draw}]
				
				\node[fullset,anchor=west] at (0,0) (A) {\normalsize $75$};
				\node[above=of A.north,indication] (ATXT) {DATASET (75 DAYS)};
				\node[fullset,anchor=west] at (0,-4) (B) {};
				\foreach \x in {0,1,...,4}
				{
					\draw (B.west) +(\x,0) node[oneblock,anchor=west,draw] {\normalsize $15$};
				}
				\draw[->] (A) -- (B) node[midway,fill=white,indication] {Divide into $5$ folds of equal size (15 DAYS)};
				
				\begin{scope}[xshift=9.1cm,yshift=2cm,scale=0.5,local bounding box=rightside box]
					\foreach \y in {1,...,5}
					{
						\draw (0,0) +(1,-\y*2) node[fullset,anchor=west] {};
						\draw (0,0) +(\y,-\y*2) node[oneblock,draw,anchor=west,fill=blue_2007] {\normalsize {\bf \color{black} Test}};
						\draw (6.8,-\y*2)  node[anchor=west] {Exp. \y $\color{blue_2007} \rightarrow$ Res. \y};
					}
					\coordinate (R) at (rightside box.west);
				\end{scope}
				\draw[->] (B.east) -- +(1.75,0) node[fill=white, align=center,indication] {Run experiments\\using $5$ different\\partitions} |- (R);
				\draw[fill =blue_2007, opacity=.2] (14.15,1.5) rectangle (15.4,-3.5);
				\draw[color=blue_2007, ->] (14.75,-3.3) -- +(0,-0.95) node[fill=white, align=center,indication] {\color{black} Compute\\ \color{black} the mean} |- (12.5,-5.1);
				\draw[fill =blue_2007,opacity=.9] (10.25,-4.65)  rectangle (12.35,-5.55);
				\draw (11.3,-5.1) node[align=center,indication] {\color{black} \bf Ensemble\\[-1mm] \color{black} \bf result};
				\node[anchor=west] at (0,-6) (C) {};
		\end{tikzpicture}}
		
		\caption{Cross-validation ensemble: the complete dataset is divided into $5$ folds. We run $5$ experiments with different partitions (of test and training). Each experiment gives a result. The final result of the ensemble is the mean of the obtained results of the experiments.}
		\label{fig:cv_kfold}
	\end{figure}
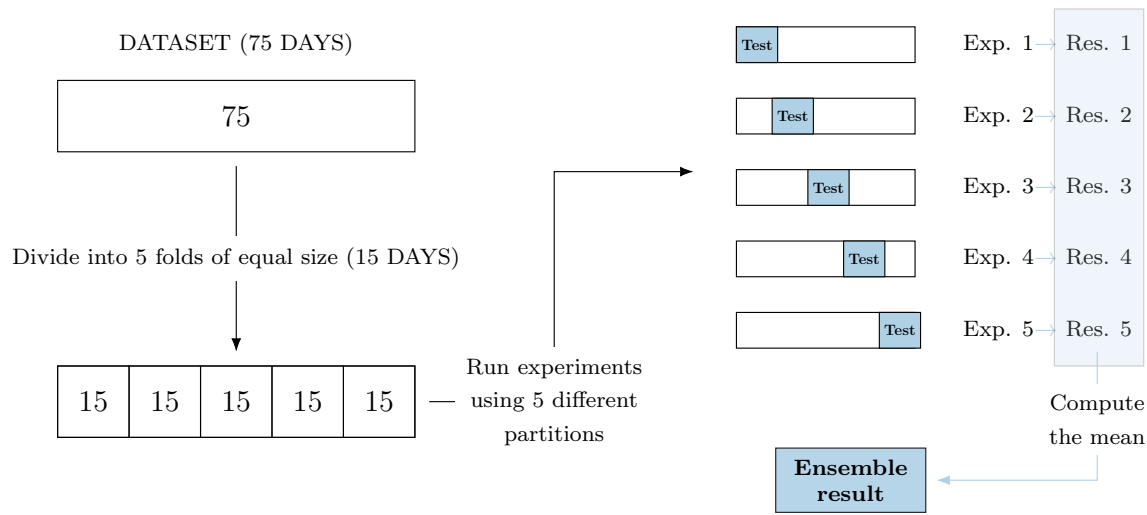

	For each fold of the cross-validation, we randomly selected $20\%$ of the days (15 days) for testing and used the remaining $80\%$ (60 days) for training the model. Importantly, the test sets form a partition of the total dataset, meaning that for each fold, the test set contains no data from the other test sets. Figure \ref{fig:cv_kfold} shows a diagram of how this partition is done. Therefore, the model is trained for each of the $5$ folds, using around $5538$ observations for training and $1369$ for testing on each one.
	
	For determining the model performance we use the Root Mean Square Error (RMSE). This metric indicates how far the model predictions are from the real data observations, being $0$ for perfect accurate predictions. RMSE provides an estimate of the number of cows for which the predictions differ from reality. Its formula is
	\begin{equation*}
		\text{RMSE} = \sqrt{ \dfrac{1}{N} \sum_{i=1}^N \big( y_i - \widehat y_i \big)^2 },
	\end{equation*}
	where $\{ y_i \}_{i=1}^N$ represents the actual data and $\{ \widehat y_i \}_{i=1}^N$ the corresponding predictions, with $N$ being the test dataset length. Note that when we report this error metric for the entire dataset, it is calculated as the mean of the values across all cross-validation folds.
	
	In the rest of the section we detail the methods used for model development and validation. It presents the performance metrics and insights from the machine learning models applied to predict the number of cows seeking shade in response to different environmental conditions. Each of these models offers different advantages in terms of accuracy and interpretability (see Section \ref{sec:intro}), and we will analyze their results to determine the most appropriate approach for this problem.
	More specifically, we begin examining the Decision Tree, a highly interpretable model that provides valuable insights into the key factors that determine cow behavior. We then move on to Random Forests, which combine multiple Decision Trees to improve prediction accuracy and reduce overfitting. Finally, we will discuss the performance of Neural Networks, which offer a more complex and flexible framework for capturing nonlinear relationships between variables but with less interpretability.

	\subsection{Soft Computing Decision Tree Algorithm}
	
\begin{figure}[h]
		\begin{forest}
			for tree={
				draw,
				rounded corners,
				edge={->,>=stealth},
				align=center,
				l sep+=25pt,
				s sep+=10pt,
				inner sep=5pt,
				minimum height=3mm,
				minimum width=12mm,
				text centered,
				font=\scriptsize,
				grow=south
			},
			[{\color{black} \textbf{THI $\leq$ 80.05}}, fill=blue_2007, 
			[{\color{black} \textbf{time $\leq$ 08:59:31}}, fill=blue_1687, edge={->,>=stealth, draw=green}, edge label={node[midway,above left, fill=green!30!white,text=black]{True}}, 
			[{\color{black} \textbf{THI accum $\leq$ 53.38}}, fill=blue_2456, edge={->,>=stealth, draw=green},
			[{\color{black} \textbf{32.91} \\ $(3.6\%)$}, fill=blue_3441, edge={->,>=stealth, draw=green}]   
			[{\color{black} \textbf{24.78} \\ $(10.9\%)$}, fill=blue_2063, edge={->,>=stealth, draw=red}] 
			]
			[{\color{black} \textbf{THI $\leq$ 75.70}}, fill=blue_754, edge={->,>=stealth, draw=red}, 
			[{\color{black} \textbf{8.79} \\ $(48.1\%)$}, fill=blue_651, edge={->,>=stealth, draw=green}]   
			[{\color{black} \textbf{20.52} \\ $(23.7\%)$}, fill=blue_1766, edge={->,>=stealth, draw=red}] 
			]
			]
			[{\color{black} \textbf{time $\leq$ 17:29:15}}, fill=blue_3858, edge={->,>=stealth, draw=red}, edge label={node[midway,above right, fill=red!30!white,text=black]{False}},
			[{\color{black} \textbf{THI accum $\leq 75.57$}}, fill=blue_4385, edge={->,>=stealth, draw=green},
			[{\color{black} \textbf{33.97} \\ $(7.7\%)$}, fill=blue_3441, edge={->,>=stealth, draw=green}]   
			[{\color{black} \textbf{56.14} \\ $(3.5\%)$}, fill=blue_4725, edge={->,>=stealth, draw=red}]  
			]
			[{\color{black} \textbf{time $\leq$ 18:38:45}}, fill=blue_1726, edge={->,>=stealth, draw=red},
			[{\color{black} \textbf{24.53} \\ $(1.6\%)$}, fill=blue_2456, edge={->,>=stealth, draw=green}] 
			[{\color{black} \textbf{6.00} \\ $(0.8\%)$}, fill=blue_651, edge={->,>=stealth, draw=red}]  
			]
			]
			]
		\end{forest}
		\caption{Decision Tree. The color represents the number of cows in the shade: the more intense the color, the more cows are in the shade that meet these conditions. The values in the terminal nodes are the predictions for the number of cows in the shade, and the percentages indicate the proportion of samples that meet the condition (there are a total of $5538$ samples).}
		\label{fig:tree}
	\end{figure}
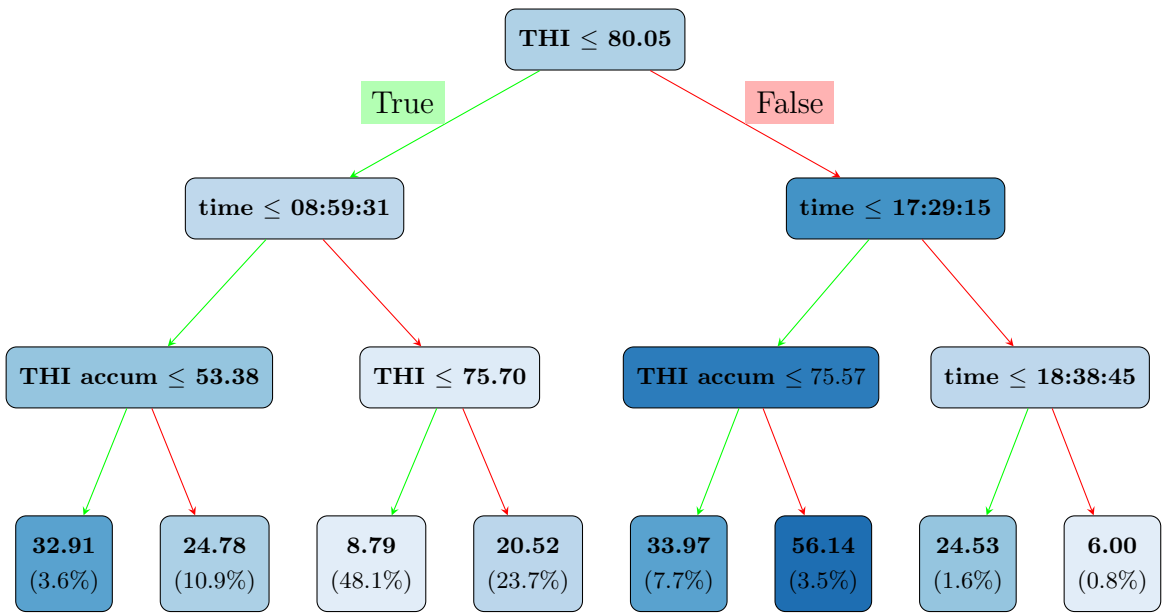

	A {\bf Decision Tree} is a type of supervised learning algorithm used for both classification and regression tasks. It works by splitting a dataset into smaller subsets according to the possible outcomes that may occur depending on decisions, as shown for example in Figure \ref{fig:tree}. The tree starts at the root node  (representing the entire dataset and the first attribute/feature) and splits the data into subsets based on an attribute that maximizes a specific criterion.

	Let us explain how the algorithm divides each node (\cite{quinlan1993c4,safavian1991survey}).
	Assume that a particular $N$ node has some observations.
	The predicted value of the model for that node consists of the mean $\overline{y}_N$ of the values of the observations in it.
	Then, the training squared error can be computed as
	\begin{equation*}
		E_N = \dfrac{1}{|N|} \sum_{x_i \in N} (x_i - \overline{y}_N)^2 = \sigma_N^2,
	\end{equation*}
	that is, the variance (here $|N|$ is the number of elements on $N$).
	However, many points with different values can be in the same node cause a huge error, so it will be split into two groups (nodes), $N = N_1 \cup N_2$, where
	\begin{equation*}
		N_1 =\Big\{x_i \in N : x_i^j \leq t\Big\}, \quad N_2 = \Big\{x_i \in N : x_i^j > t\Big\}.
	\end{equation*}
	
	To achieve this, a variable $j$ and a threshold $t$ must be selected such that the resulting partition creates two new nodes ($N_1$ and $N_2$) with the lowest possible error, minimizing $E_{N_1} + E_{N_2}$. This process is repeated for each node that contains more observations than a predefined threshold (in our case 2), or when the node reaches a {\it depth} greater than a set value, which represents the number of splits from the root node to the current one. The final nodes, which can no longer be split, are called leaves.
	
	Once the tree is trained, that is, all the conditions and thresholds are known, its prediction of a new observation $x$ is computed as follows.
	The observation starts from the root node and follows the branches according to the conditions satisfied until it arrives at a leaf $L$.
	The predicted value for $x$ is $\overline{y}_L$, the mean of all training observations in the node $L$.
	
	The deeper the tree and the smaller the allowed nodes (in terms of the minimum number of observations required at each node), the smaller the error in the training set.
	In the extreme case where each leaf consists of only one observation, the training error will be $0$, but this does not mean that the error will be lower when predicting new observations (for example, on the test set). 
	This phenomenon is known as \emph{overfitting}, and  we will see how the selection of hyperparameters is done to allow the model to be flexible enough to fit the data without being overfitted.

	In our particular case the Decision Tree model is particularly well-suited for understanding how individual features influence cow behavior, as it visually represents decision rules in a hierarchical structure. This makes it easy to identify the thresholds and conditions under which cows are most likely to seek shade.

	In the following, we present the structure of the Decision Tree model applied to our dataset. Figure \ref{fig:tree} shows a tree with a depth of $3$ for illustrative purposes, offering a clear representation of the key factors involved (current THI, time of day and THI accumulation). Indeed, as can be seen, the main factor at the root of the tree is THI, with a threshold value of $80.05$. This indicates that when THI exceeds this value, cow behavior changes significantly, with more cows seeking shade.

	However, as shown in Table \ref{fig:tree_errors}, the optimal depth for the Decision Tree model is $5$, yielding an RMSE of $16.027$. This depth level allows the model to capture the complexity of the relationships between THI, time, and cow behavior without overfitting the data. Interestingly, as the depth increases above $5$, the error also begins to increase. This suggests that the model begins to overfit, capturing noise instead of meaningful patterns. Overfitting is a common problem in Decision Trees, especially when the depth is allowed to grow too large, as the model becomes too specific to the training data, losing generalizability. 
	Of course, although deeper models can incorporate more variables, potentially including THI from the previous night, the lack of relevance in the $3$-depth model indicates that immediate environmental conditions play a much more important role in determining cow movements to shaded areas.

	\begin{table}[h]
		\centering	\begin{tabular}{lccccccc}
			\toprule
			\textbf{Depth} & $1$ & $3$ & \color{blue} $5$ & $10$ & $15$ & $25$ & $50$ \\
			\textbf{RMSE} & $17.994$ & $16.711$ & \color{blue} $16.027$ & $18.764$ & $19.941$ & $20.235$ &  $20.219$ \\
			\bottomrule \smallskip
		\end{tabular}
		\caption{Decision Tree errors (RMSE) for different tree depths, ranging from $1$ to $50$. The model with the lowest error is highlighted in blue.}
		\label{fig:tree_errors}
	\end{table}

	Although the Decision Tree provides a clear and interpretable model, it may be interesting to explore other methods that capture more complex interactions between variables. This is the purpose of the next type of algorithms.

	\subsection{Soft Computing Random Forest Algorithm}
	
	As the performance of a Decision Tree is limited, an ensemble of many of them can be considered to form a {\bf Random Forest}---which also works for classification and regression tasks---. This method is particularly effective for improving accuracy while mitigating overfitting. 
	The Random Forest is an ensemble of multiple Decision Trees combining the predictions of several base estimators to improve robustness and accuracy. It provides estimates of feature importance, which can help in understanding the underlying structure of the data (see \cite{breiman2001random,liaw2002classification}). This model is less interpretable compared to a single Decision Tree due to the aggregation of multiple trees and can also be slower to train compared to simpler models, especially with large datasets. 
	
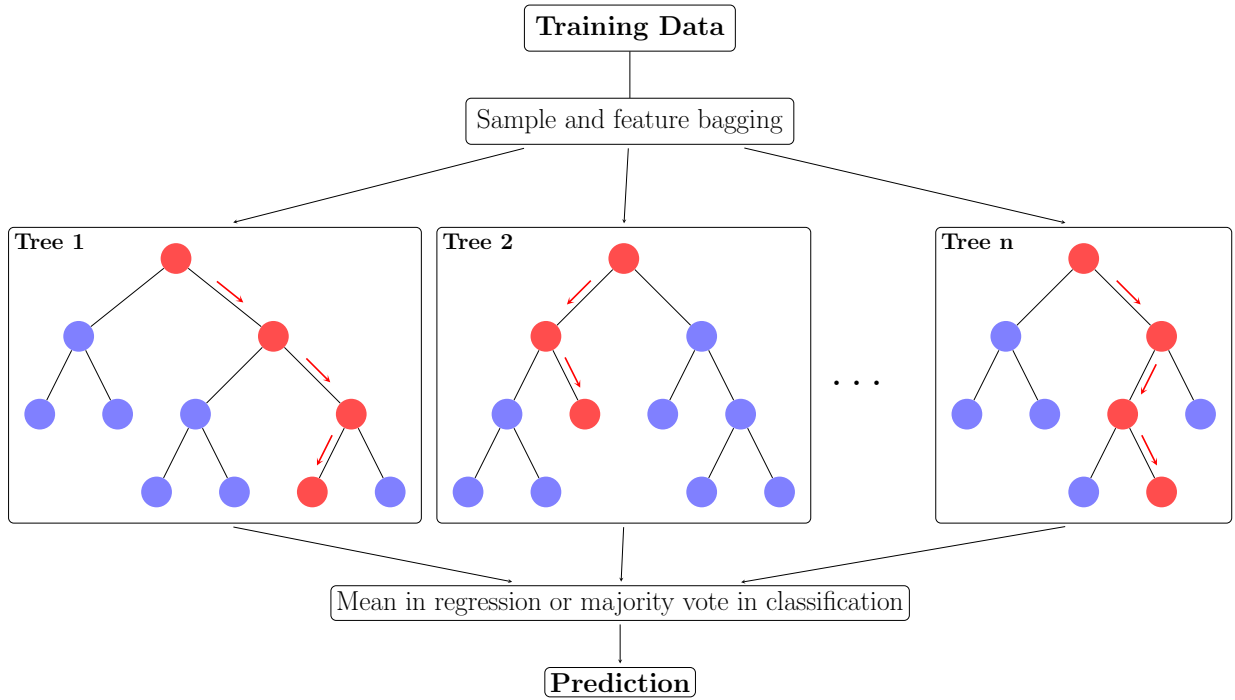
\begin{figure}[h]
		\tikzset{
			font=\Large,
			red arrow/.style={
				midway,red,sloped,fill, minimum height=3cm, single arrow, single arrow head extend=.5cm, single arrow head indent=.25cm,xscale=0.3,yscale=0.15,
				allow upside down
			},
			black arrow/.style 2 args={-stealth, shorten >=#1, shorten <=#2},
			black arrow/.default={1mm}{1mm},
			tree box/.style={draw, rounded corners, inner sep=1em},
			node box/.style={white, draw=black, text=black, rectangle, rounded corners},
		}

		\scalebox{0.5}{\begin{forest}
				for tree={l sep=3em, s sep=3em, anchor=center, inner sep=0.7em, fill=blue!50, circle, where level=2{no edge}{}}
				[
				\textbf{\LARGE Training Data}, node box
				[\LARGE Sample and feature bagging, node box, alias=bagging, above=4em
				[,red!70,alias=a1[[,alias=a2][]][,red!70,edge label={node[above=1ex,red arrow]{}}[[][]][,red!70,edge label={node[above=1ex,red arrow]{}}[,red!70,edge label={node[below=1ex,red arrow]{}}][,alias=a3]]]]
				[,red!70,alias=b1[,red!70,edge label={node[below=1ex,red arrow]{}}[[,alias=b2][]][,red!70,edge label={node[above=1ex,red arrow]{}}]][[][[][,alias=b3]]]]
				[~~$\dots$~,scale=2,no edge,fill=none,yshift=-4em]
				[,red!70,alias=c1[[,alias=c2][]][,red!70,edge label={node[above=1ex,red arrow]{}}[,red!70,edge label={node[above=1ex,red arrow]{}}[,alias=c3][,red!70,edge label={node[above=1ex,red arrow]{}}]][,alias=c4]]]]
				]
				\node[tree box, fit=(a1)(a2)(a3)](t1){};
				\node[tree box, fit=(b1)(b2)(b3)](t2){};
				\node[tree box, fit=(c1)(c2)(c3)(c4)](tn){};
				\node[below right=0.5em, inner sep=0pt] at (t1.north west) {\textbf{Tree 1}};
				\node[below right=0.5em, inner sep=0pt] at (t2.north west) {\textbf{Tree 2}};
				\node[below right=0.5em, inner sep=0pt] at (tn.north west) {\textbf{Tree $\mathbf{n}$}};
				\path (t1.south west)--(tn.south east) node[midway,below=4em, node box] (mean) {\LARGE Mean in regression or majority vote in classification};
				\node[below=3em of mean, node box] (pred) {\textbf{\LARGE Prediction}};
				\draw[black arrow={5mm}{4mm}] (bagging) -- (t1.north);
				\draw[black arrow] (bagging) -- (t2.north);
				\draw[black arrow={5mm}{4mm}] (bagging) -- (tn.north);
				\draw[black arrow={5mm}{5mm}] (t1.south) -- (mean);
				\draw[black arrow] (t2.south) -- (mean);
				\draw[black arrow={5mm}{5mm}] (tn.south) -- (mean);
				\draw[black arrow] (mean) -- (pred);
		\end{forest}}
		\caption{Random Forest working example.}
		\label{fig:RF}
	\end{figure}

	The algorithm works as follows:  given an observation $x$, the output of the Random Forest is given by the mean of the prediction of all Decision Trees in the case of regression or the majority vote in classification (see Figure \ref{fig:RF}).
	
	To avoid having the same Tree each time, which would have no improvement when averaging them, some randomness is intentionally introduced on each one, which is usually both included in the training set and the variables used.
	A fixed number of variables are randomly selected for each tree (three out of four, in our case).
	Moreover, not all training set is considered on each Tree a sample of the training dataset allowing duplicates.
	This process is known as \emph{bootstrapping}.
	
	\begin{figure}[h]
		\centering
		\includegraphics[width=0.7\linewidth]{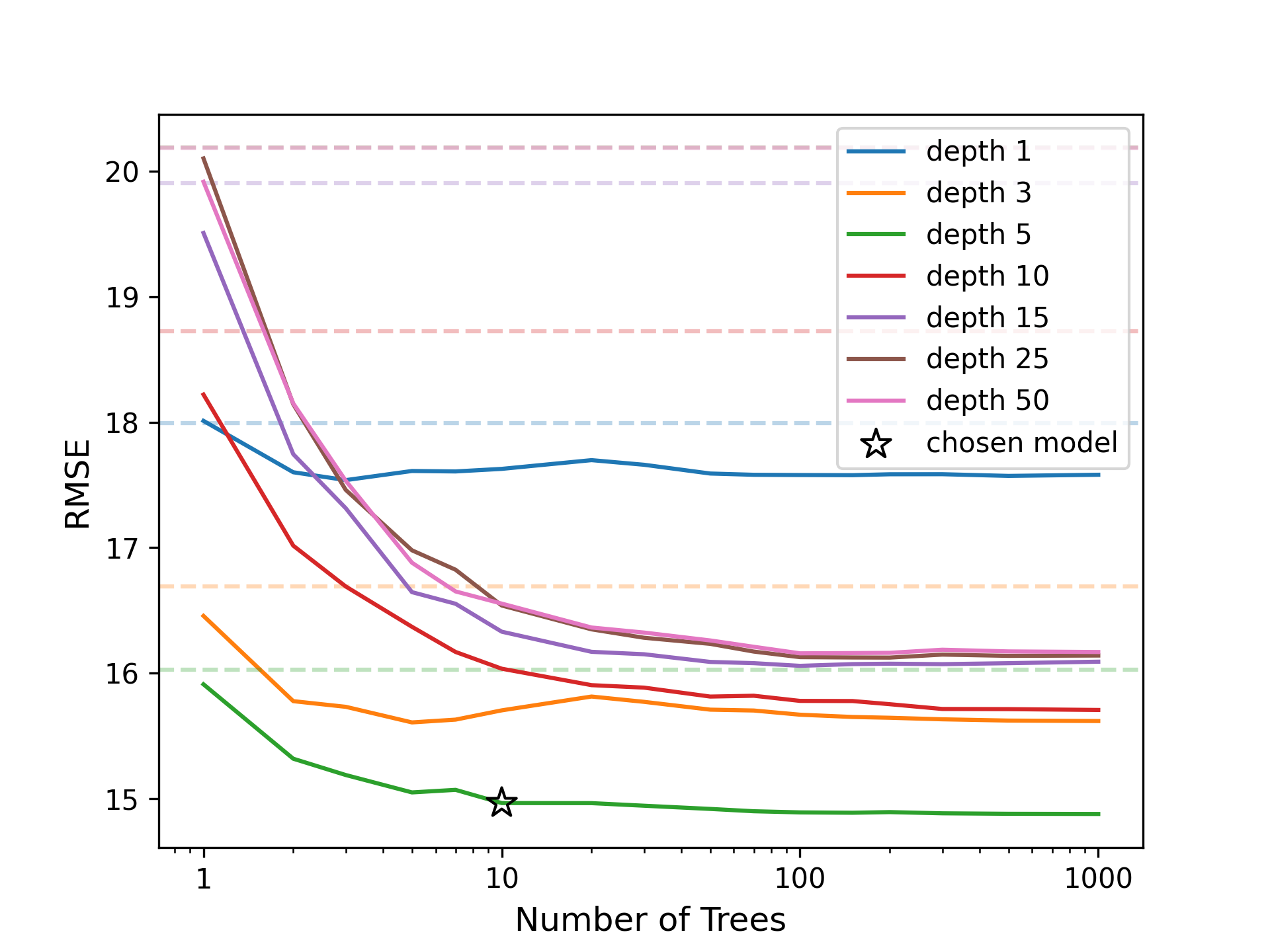}
		\caption{Random Forest errors (RMSE) for varying numbers of trees  (ranging from $1$ to $1000$) and depths (ranging from $1$ to $50$). The dashed lines in lighter colors represent the errors of single Decision Trees with corresponding depths, used as a reference for comparison with the Random Forest model. The chosen model, marked with a star, balances accuracy and computational efficiency.}
		
		\label{fig:rmserf5cv}
	\end{figure}
	
	The performance of the Random Forest model is evaluated in comparison to single Decision Trees, focusing on its error reduction capabilities as the number of trees increases and in relation to tree depth. In Figure \ref{fig:rmserf5cv}, we observe that the model with a tree depth of $5$ (see the green line) achieves the best fit for our dataset, consistent with the optimal depth observed in the Decision Tree model. As the number of trees increases, the error (measured by RMSE) decreases, demonstrating the benefit of using an ensemble of trees. However, we opted to use a model with $10$ trees, which offers a reasonable balance between accuracy and interpretability, achieving an RMSE of $14.965$, which is only around $0.08$ higher than the model with $1000$ trees.

	The trade-off between the number of trees and model interpretability is crucial. While more trees generally improve accuracy, especially in complex, high-dimensional datasets, the added complexity can make it harder to interpret the model’s behavior. In our case, using $10$ trees allows us to retain a high level of interpretability while achieving near-optimal performance.

	\subsection{Soft Computing Neural Networks Algorithm}
	
\begin{figure}[h]
		\centering
		\begin{tikzpicture}[scale=0.9]
			\definecolor{inputcolor}{RGB}{173,216,230}  
			\definecolor{hiddencolor}{RGB}{144,238,144} 
			\definecolor{outputcolor}{RGB}{255,182,193} 
			
			
			\node at (-3, 1.5) {\textbf{$x$}};
			
			\node[circle, draw, fill=inputcolor] (in1) at (0, 2.5) {};
			\node[circle, draw, fill=inputcolor] (in2) at (0, 1.5) {};
			\node[circle, draw, fill=inputcolor] (in3) at (0, 0.5) {};
			
			\node at (0, -1) {Input Layer};
			
			\node[circle, draw, fill=hiddencolor] (h11) at (3, 3) {};
			\node[circle, draw, fill=hiddencolor] (h12) at (3, 2) {};
			\node[circle, draw, fill=hiddencolor] (h13) at (3, 1) {};
			\node[circle, draw, fill=hiddencolor] (h14) at (3, 0) {};
			
			\node[circle, draw, fill=hiddencolor] (h21) at (6, 3) {};
			\node[circle, draw, fill=hiddencolor] (h22) at (6, 2) {};
			\node[circle, draw, fill=hiddencolor] (h23) at (6, 1) {};
			\node[circle, draw, fill=hiddencolor] (h24) at (6, 0) {};
			
			\node at (4.5, -1) {Hidden Layers};
			
			\node[circle, draw, fill=outputcolor] (out) at (9, 1.5) {};
			
			\node at (9, -1) {Output Layer};
			
			\node at (11.8, 1.5) {\textbf{$\widehat{y}$}};
			
			\draw[->] (-2.2, 1.5) -- (in1);
			\draw[->] (-2.2, 1.5) -- (in2);
			\draw[->] (-2.2, 1.5) -- (in3);
			
			\foreach \i in {1,2,3}
			\foreach \j in {11,12,13,14}
			\draw (in\i) -- (h\j);
			
			\foreach \i in {11,12,13,14}
			\foreach \j in {21,22,23,24}
			\draw (h\i) -- (h\j);
			
			\foreach \i in {21,22,23,24}
			\draw (h\i) -- (out);
			
			\draw[->] (out) -- (11.2, 1.5);
		\end{tikzpicture}
		\caption{Scheme of an example of a (Fully Connected) Neural Network. The input $x$ represents the input data, while $\widehat{y}$ denotes the model's prediction. The network consists of an input layer, multiple hidden layers, and an output layer.}
		\label{fig:FCNN}
	\end{figure}
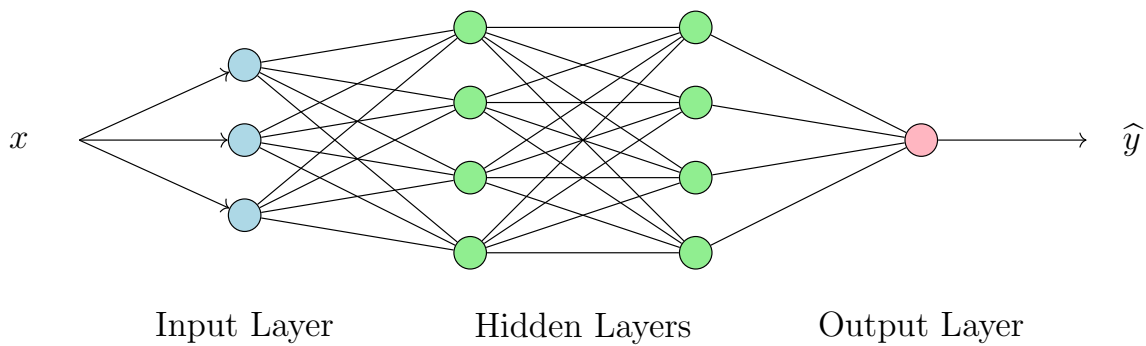

	A {\bf Neural Network} is a model inspired by the human brain's structure and function. It consists of interconnected layers of nodes (called neurons), as shown in the Figure \ref{fig:FCNN}, where each connection has a weight that adjusts as learning progresses.  It can be used to identify patterns and relationships in data through a  training process but also for classification and regression. During training, the network learns by adjusting weights based on the errors of its predictions compared to known outcomes (see \cite{goodfellow2016deep,lecun2015deep}). Each layer consists of a linear transformation and a composition with a nonlinear function.
	The number of layers, dimensions of each one and the nonlinear functions used on each are fixed as hyperparameters of the model, while the weights of the linear transformation are optimized on the training.
	
	During training, backpropagation is used to adjust the network weights. This process consists of calculating the model prediction, $\widehat{y}$, for each observation $x$ in the training dataset and comparing it to the true target, $y$. If $\widehat{y}$ does not coincide with $y$, the weights are updated to reduce the prediction error. This is achieved using a gradient-based optimization algorithm that minimizes the squared error $(y - \widehat{y})^2$ (SGD or Adam algorithm).
	Weight updates are propagated backward through the network, layer by layer. Training is repeated for all observations in the training dataset during a fixed number of times, called \emph{epochs}, or when the error in some test dataset attains a minimum (early stopping parameter).
	For the best performance of the (Fully Connected) Neural Network (FCNN) model and to avoid variables with larger scales having more influence on the predictions, all variables are scaled so that their mean is $0$ and their standard deviation is $1$.
	
	Neural Networks are powerful models capable of capturing complex, non-linear relationships between inputs and outputs. By stacking multiple layers of neurons, these networks can approximate intricate patterns in the data effectively. In the following sections, we will assess the performance of different configurations of the Neural Network. Our focus will be on how factors such as learning rate, the number of neurons per layer, and the total number of layers influence the predictive accuracy of the model.
	
	There are several activation functions available for use in Neural Networks (\cite{survey_fact, fact}). However, for our FCNN implementation, we have selected the Rectified Linear Unit activation function for hidden layers. This function, mathematically expressed as $\text{ReLU}(x)= \max(0,x)$, allows the network to efficiently model nonlinear relationships in the data. For the output layer, we use a linear activation function defined as $\text{Linear}(x)=x$. This ensures that the output can take any real value, which is suitable for our prediction task.
	
	The primary challenge when working with Neural Networks lies in determining the optimal architecture—balancing depth (number of layers), width (number of neurons per layer), and learning rate—. While deeper and wider networks have the potential to capture more intricate patterns, they also increase the risk of overfitting and may require significantly more computational resources. Hence, selecting the right configuration is critical to achieving both accuracy and efficiency.
	To rigorously identify the architecture, 45 different neural networks have been tested with all the combination of these hyperparameters: 1, 3 or 5 layers with 4, 16, 64, 256 or 1024 neurons each and a learning rate of $10^{-2}$, $10^{-3}$ or $10^{-4}$.
	
	\begin{table}[h]
		\centering	\begin{tabular}{lrrrrr}
			\toprule
			&      \textbf{lr} &  \textbf{neurons} &  \textbf{layers} &      \textbf{Parameters} & \textbf{RMSE} \\
			\midrule
			1 &  $10^{-3}$ &       $16$ &       $5$ & $1185$ &  $14.729544$ \\
			\color{blue} 2 &  \color{blue} $10^{-3}$ &       
			\color{blue} $16$ &       \color{blue} $3$ & \color{blue} $641$ & \color{blue} $14.783720$ \\
			3 &  $10^{-4}$ &      $256$ &       $3$ & $133121$ & $14.841024$ \\
			4 &  $10^{-4}$ &       $64$ &       $5$ & $17025$ & $14.892802$ \\
			5 &  $10^{-3}$ &     $1024$ &       $1$ & $8705$ & $14.908759$ \\
			6 &  $10^{-4}$ &       $64$ &       $3$ & $8705$ & $14.983424$ \\
			7 &  $10^{-2}$ &       $64$ &       $1$ & $385$ & $15.005318$ \\
			8 &  $10^{-3}$ &      $256$ &       $1$ & $1537$ & $15.127514$ \\
			9 &  $10^{-2}$ &      $256$ &       $1$ & $1537$ & $15.414012$ \\
			10 &  $10^{-3}$ &       $64$ &       $1$ & $385$ & $15.565035$ \\
			\bottomrule \smallskip
		\end{tabular}
		\caption{Performance of the best error-based models (RMSE) with different learning rates (lr), neurons, and layers. We also show the number of parameters to optimize. In blue color best model balancing complexity (number of parameters) and RMSE.}
		\label{fig:resultsnn}
	\end{table}

	Table \ref{fig:resultsnn} shows the performance of eight Neural Network models with varying configurations. The results indicate that the most efficient model, in terms of balancing complexity and error, is Model 2, which has $3$ hidden layers with $16$ neurons in each layer and a learning rate of $10^{-3}$. This model achieves an RMSE of $14.784$ using only $641$ parameters, making it accurate and relatively simple compared to other more complex models. Below, we discuss the key factors that affect the performance of Neural Networks.
	
	One clear observation from the results is that increasing the number of \textbf{neurons per layer} does not always yield better results. For instance, Model 1, with $5$ (hidden) layers and $16$ neurons per layer, achieves an RMSE of $14.73$, comparable to Model 4, which has $64$ neurons per layer and a slightly higher error ($14.89$). This suggests that simpler architectures can perform well, avoiding excessive model intricacy.
	Additionally, models with fewer \textbf{layers}, such as Model 2 with only $3$ layers and $16$ neurons per layer, achieve competitive error values, demonstrating that adding more layers may introduce unnecessary complexity without significant performance gains.
	
	Another crucial hyperparameter that affects model performance is the \textbf{learning rate (lr)}. In our experiments, we tested different rates: $10^{-2}$, $10^{-3}$ and $10^{-4}$.
	The results reveal that higher learning rates, such as $10^{-2}$ (Model 7, RMSE $15.005$), hinder convergence, while lower rates tend to produce lower error values, especially when used with a smaller number of layers and neurons.
	
	Finally, in the five best configurations, the RMSE values remain in a narrow range between $14.7$ and $14.9$. Despite the variations in layers, neurons and learning rates, the best performing models (Models 1 and 2) show very close error values, with a difference of only $0.05$. Given these minimal differences, we selected Model 2, which has fewer \textbf{parameters} ($641$) and is therefore more computationally efficient without compromising accuracy. This balance between simplicity and performance makes it an ideal choice for practical applications requiring faster training times and lower resource consumption.

	\begin{figure}[h]
		\centering    \includegraphics[width=.5\linewidth]{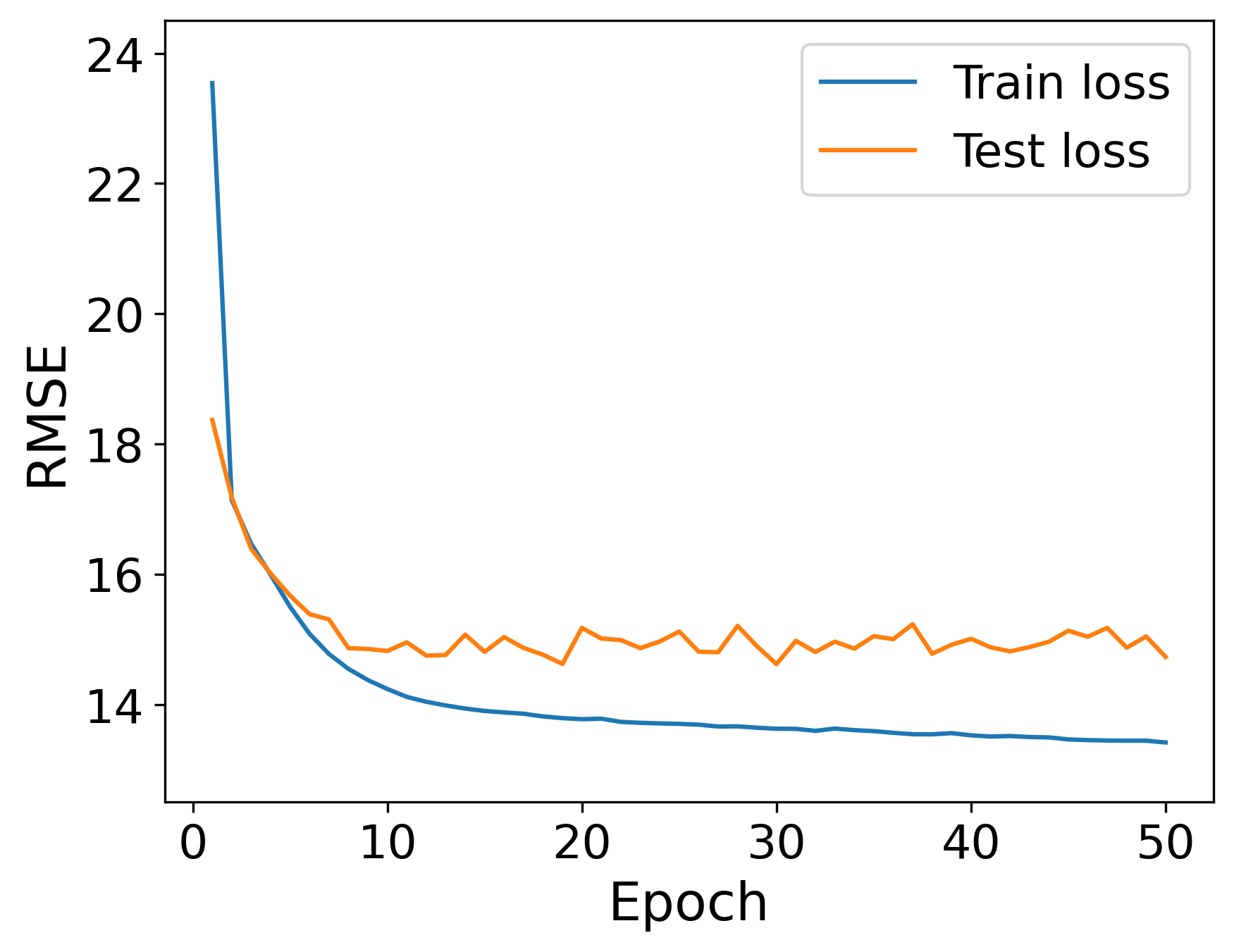}
		\caption{Evolution of RMSE for Neural Networks over $50$ training epochs.}
		\label{fig:loss_NN}
	\end{figure}

	From Figure \ref{fig:loss_NN}, it is evident that increasing the number of epochs beyond $10$ does not significantly improve the model error. After about a decade of epochs further training offers little to no advantage in terms of predictive accuracy. This insight is particularly relevant for applications requiring instant model recalculations, such as real-time systems, where the model could effectively operate with this number of epochs. This approach allows for greater efficiency in terms of training time and computational resources without significantly affecting the model's performance.
	
	\section{Results}\label{sec:results}
	
	\subsection{Global model performance}
	
	We begin by comparing the three machine learning models implemented: Decision Tree with $5$ depth, Random Forest consisting of $10$ trees all with $5$ depth and a Neural Network with $3$ hidden layers of $16$ neurons each and $10^{-3}$ learning rate. The comparison is given not only in terms of RMSE, but also in terms of interpretability and explainability (see Figure \ref{fig:exp_int}).
	
	\begin{table}[h]
		\centering
		\begin{tabular}{lcll}
			\toprule
			\textbf{Model} & \textbf{RMSE} & \textbf{Interpretability} & \textbf{Explainability} \\ \midrule
			Decision Tree & $16.03$ & Very High & High \\ 
			\color{blue} Random Forest & \color{blue} $14.97$ & \color{blue} Medium/High & \color{blue} Medium \\ 
			Neural Network & $14.78$ & Low & Low \\ \bottomrule \smallskip
		\end{tabular}
		\caption{Comparison of final model using three criteria: RMSE, Interpretability and Explainability.}
		\label{tab:model_comparison}
	\end{table}
	
	On the one hand, the Decision Tree model has a RMSE of $16.03$, which is the highest among the models compared, meaning it is less accurate for prediction. However in terms of Interpretability, the Decision Tree scores {\it Very High} (that  means the model is easy to understand), as its structure is simple and intuitive, resembling a flowchart. In a similar way the  Explainability  of the Decision Tree is rated as {\it High}, meaning that the decision-making process of the model can be clearly explained, making it easier to trace the reasoning behind predictions.
	
	The Random Forest model has a slightly lower RMSE of $14.97$, indicating better accuracy compared to the Decision Tree.
	Its  Interpretability is rated {\it Medium/High}. Although more complex than a single Decision Tree due to the ensemble of trees, it still maintains some interpretability because individual trees can be analyzed. The  Explainability  of the Random Forest model is rated as {\it Medium}, as it is harder to fully explain how multiple trees work together in the ensemble, but some level of explanation is still possible.
	
	Finally, the Neural Network model has the lowest RMSE of $14.78$, making it the most accurate of the three models in terms of predictions.
	However, the Interpretability of this model is {\it Low}, meaning it is difficult to understand how the model works, due to its complex structure with layers of interconnected neurons. Similarly, the Explainability  is rated {\it Low}, as it is challenging to explain how the model arrives at specific predictions, making it a \textit{black box} in many cases.
	
	This comparison highlights a trade-off between accuracy (RMSE) and interpretability/explainability, where models with better accuracy, like the Neural Network, are harder to interpret and explain. Conversely, the Decision Tree offers higher interpretability and explainability but with slightly lower accuracy. Thus Random Forest would be an intermediate model in terms of the error and interpretability/explainability.
	
	\subsection{Error distribution analysis}

\begin{figure}[h]
		\centering
		\includegraphics[scale=.35]{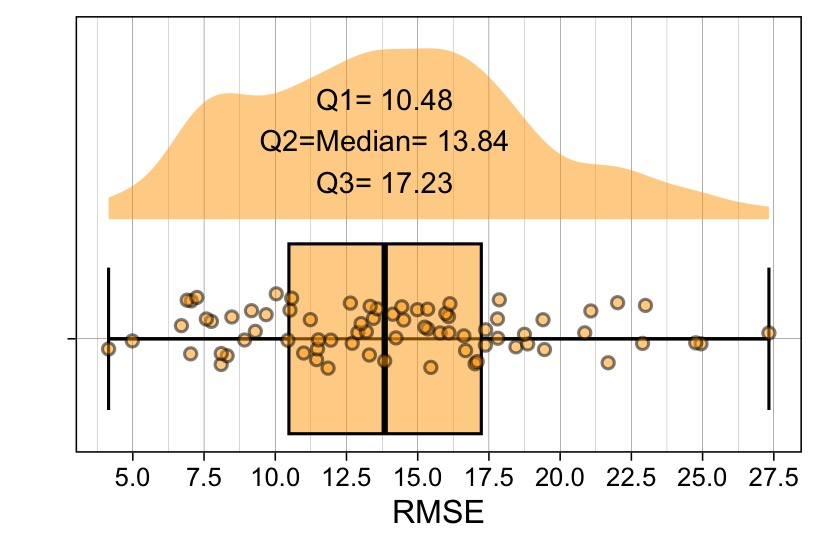}
		\caption{Raincloud plot representing the distribution of Random Forest errors (RMSE) calculated for each day across each cross-validation partitions. The box shows the interquartile range (IQR), with the median error represented by the central line within the box, which is $13.84$.
			The whiskers extend to the minimum and maximum RMSE values.
			Distribution of the days and the first ($Q1$) and third ($Q3$) quartiles are also given.}
		\label{fig:preds_RF_boxplot}
	\end{figure}

	Now, we compare the real and predicted values for the number of animals seeking shade over the span of 75 days. By analyzing the raincloud plot shown in Figure \ref{fig:preds_RF_boxplot}, we can draw several conclusions about the model’s performance in terms of RMSE values.
	
	First, the model’s predictions are generally consistent, as indicated by a median RMSE of $13.84$. This relatively low median error suggests that the model performs accurately on average. The interquartile range (IQR), from $Q1$ at $10.48$ to $Q3$ at $17.23$, shows that the middle $50\%$ of RMSE values are concentrated within a fairly narrow range. This concentration suggests that most predictions fall within a predictable error margin, which is desirable in applications requiring reliability and stability.

	Second, the plot reveals a few RMSE values extending toward the extremes (and above 25). This indicates that, in certain cases, the model performs less accurately. These outliers may correspond to specific scenarios or data points where the model struggles to generalize, possibly due to variations in environmental conditions or unmodeled factors. Addressing these cases could involve incorporating additional features or refining the model to improve its generalizability.
	
	\subsection{Case studies}
	
	Hereafter, we present the real and predicted values for the number of animals in the shade over the course of the day corresponding to the first quartil in terms of that day's RMSE (August 18th, 2023), by using our Random Forest algorithm. We also includes information about THI (accumulated day, mean night and current).

	\begin{figure}[h]
		\centering
		\includegraphics[scale=.7]{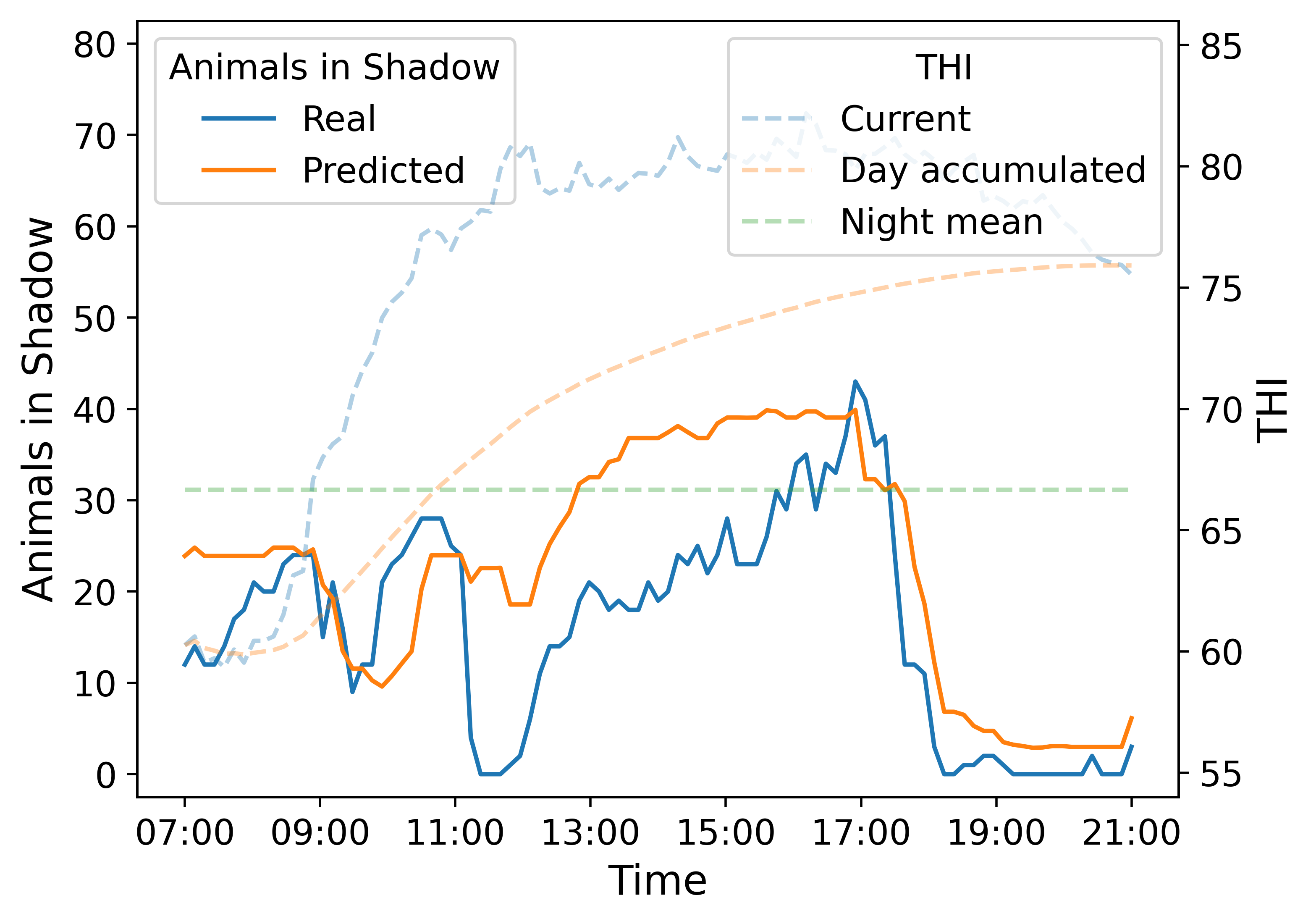}
		\caption{Random Forest predictions for August 18th, 2023, the day corresponding to $Q1$.
			The left $Y$-axis (with continuous lines) represents the number of  animals in the shade (ranging from $0$ to $80$), the right $Y$-axis (with dashed lines) indicates the value of the THI (ranging from $55$ to $85$) and finally  the $X$-axis displays the time of day (ranging from 07:00 in the morning to 21:00 in the evening). The blue line represents the real data, and the orange line shows the predicted values. Additionally, the dashed blue line represents the current THI, the dashed orange line indicates the accumulated THI throughout the day, and the green dashed line represents the average nighttime THI of the previous day.}
		\label{fig:preds_RF1}
	\end{figure}

	As illustrated in Figure \ref{fig:preds_RF1}, early in the day, from 07:00 until around 11:00, the number of animals in the shade is low, which corresponds with lower THI values. Both real and predicted values are similar during this period. Between 11:00 and 17:00, as the Current THI increases sharply, there is a remarkable increase in the number of animals seeking shade. This increase is evident in both the real and predicted data, although the predicted values (orange line) show a smoother and less variable trend. After 17:00, as THI decreases, the number of animals in shade also drops significantly, and both real and predicted values approach zero towards the end of the day. The predictions of the Random Forest model (orange line) follow the general trend of the real data (blue line) reasonably well. However, there are some discrepancies, especially towards midday and early afternoon (from 13:00 to 17:00), where the model slightly overestimates the number of animals in shade. On the other hand the THI accumulated during the day (dashed orange line) increases throughout the day, reflecting the cumulative heat stress experienced over time. This cumulative THI could have a prolonged impact on animal comfort, contributing to the increased use of shaded areas as animals try to avoid heat stress.
	
	\begin{figure}[h]
		\begin{subfigure}[b]{0.5\textwidth}
			\includegraphics[scale=.44]{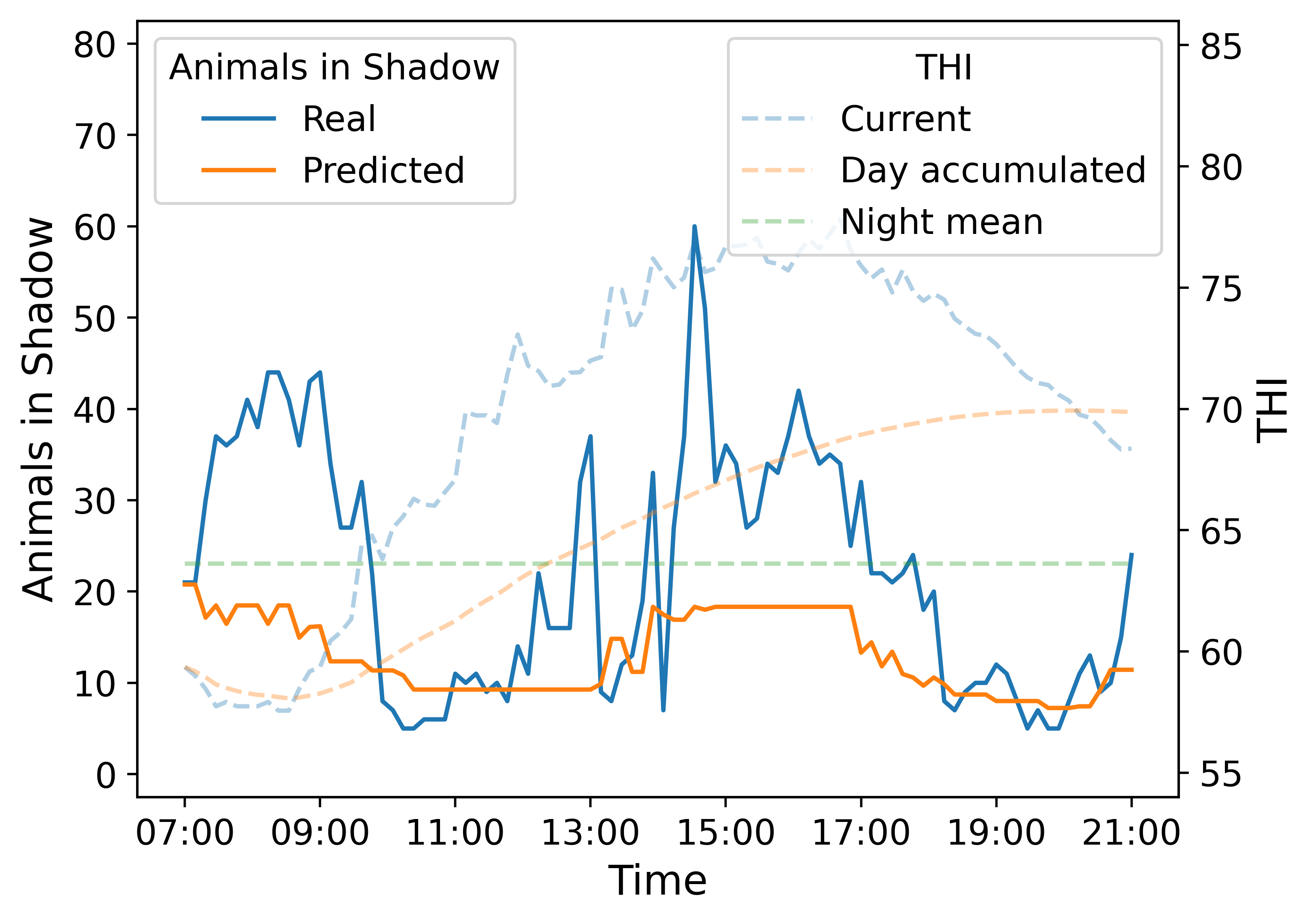}
			\caption{Model predictions for September 9th, 2023 ($Q2$).}
			\label{fig:preds_RF2_a}
		\end{subfigure} 
		\begin{subfigure}[b]{0.45\textwidth}
			\includegraphics[scale=.44]{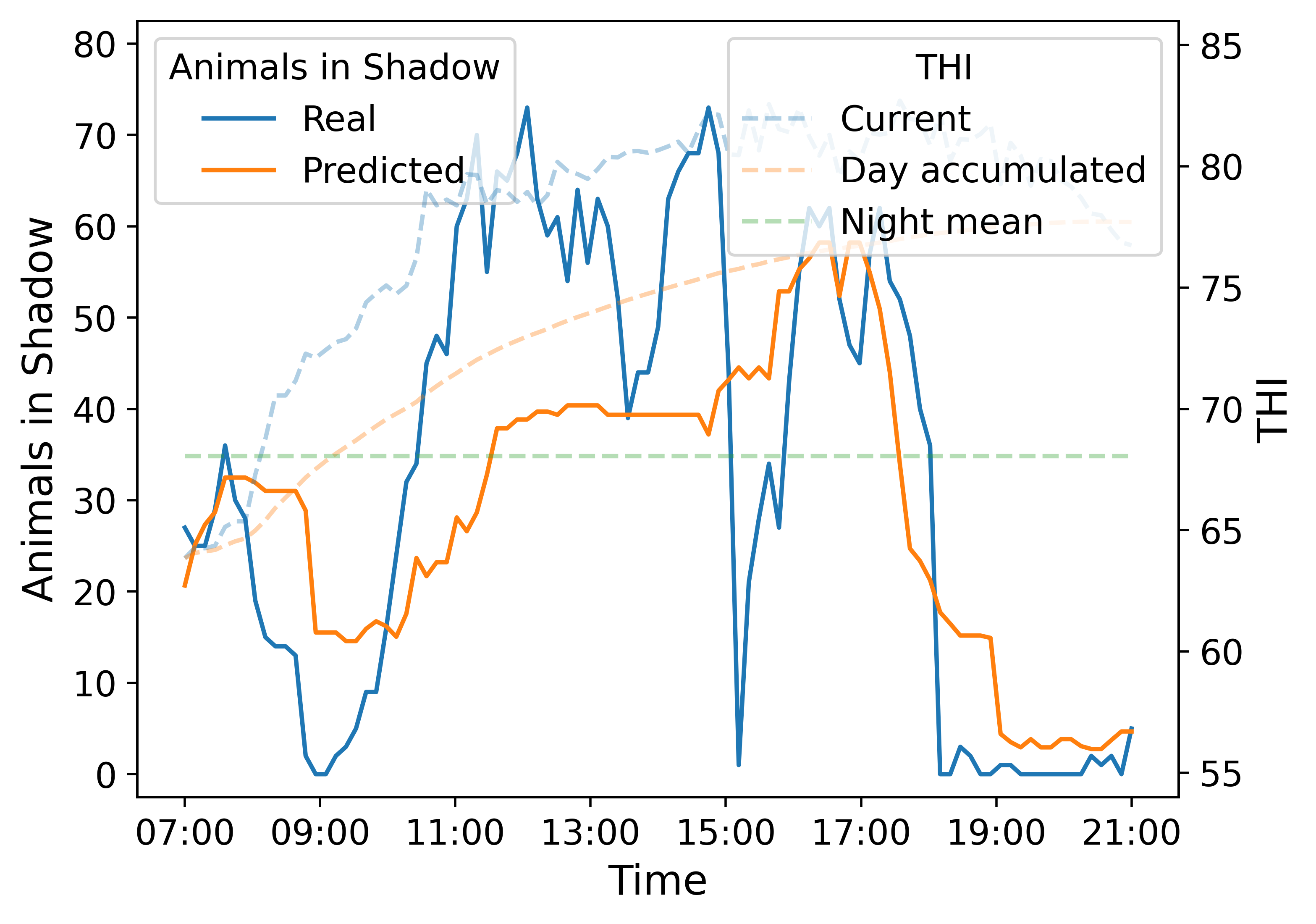}
			\caption{Model predictions for July 20th, 2023 ($Q3$).}
			\label{fig:preds_RF2_b}
		\end{subfigure}
		\caption{Random Forest predictions for specific dates corresponding to the median (\subref{fig:preds_RF2_a}) and the third quartiles (\subref{fig:preds_RF2_b}). These figures follow the same format as Figure \ref{fig:preds_RF1}.}
		\label{fig:preds_RF2}
	\end{figure}
	
	Finally, as we can also see in Figure \ref{fig:preds_RF2}, the prediction model in the three plots ($Q1$, $Q2$ and $Q3$) follows the general trend but, as expected, the prediction overlooks short-term fluctuations. Particularly, it predicts when more animals move into the shade and when they leave with an accuracy of less than one hour in most  cases. However, it cannot find the exact number of cows in the shadow area. This fact, may indicate that other factors affect to the animals decisions, but the time, THI and accumulate THI at day and night explain most of its behavior.
	
	\section{Discussion}\label{sec:discussion}
	
	In this work, we have analyzed the performance of three soft computing Machine Learning models---Decision Trees, Random Forests, and Neural Networks---to predict the number of cows seeking shade as a response to varying environmental conditions. Using data from a farm in Titaguas, Valencia, the research aimed to determine which model could best predict shade-seeking behavior in response to the Temperature-Humidity Index (THI), a key indicator of thermal stress in cattle, while also exploring the capabilities and limitations of these models for livestock management under heat stress conditions.
	
	The results have shown that each model has distinct strengths. Neural Networks provided the highest accuracy, with a root mean square error (RMSE) of $14.78$, followed closely by the Random Forest at $14.97$ and the Decision Tree at $16.03$. However, model performance was not evaluated solely based on accuracy. Interpretability and explainability were also central to the evaluation, especially for practical applications in farm management. Although the Decision Tree model is the most interpretable (given its flowchart structure that allows direct analysis of how different variables influence predictions), Random Forests, while more complex, retain some interpretability as individual trees can be analyzed to understand decision pathways. In contrast, Neural Networks, despite their high accuracy, were the least interpretable due to their multi-layered structure, often referred to as a \textit{black box} in machine learning.
	
	Taking these factors into account, Random Forests has been chosen as the optimal model, offering a balance between accuracy, interpretability, and explainability. It has effectively captured overall trends in cow movement patterns, accurately predicting when cows would seek or leave shaded areas within an hour’s precision in most instances. This precision is important in real-world applications, where timely interventions can help mitigate the effects of heat stress.
	
	The performance gap between the Random Forest model and the Neural Network is small (around $0.2$ RMSE), yet the tree-based ensemble is far easier to interpret and inherently robust to noisy or missing inputs---an essential property in commercial farms where cameras may be hidden and low-cost sensors drift over time---. This robustness makes the model an ideal upstream component for a {\bf hybrid soft-computing controller}: its predicted shade-seeking count can feed a fuzzy-logic rule base that activates fans, sprinklers or retractable awnings according to linguistic rules such as: "{\tt if} THI is high {\tt and} Predicted Shade is many, {\tt then}  increase airflow". These neuro-fuzzy or RF-fuzzy combinations retain the data-driven accuracy of machine learning, while adding transparent, expert-defined control actions, providing a practical path to farm climate management.
	
	In this article, several variables influencing shade-seeking behavior have been identified: in particular the time of day, current THI, and cumulative THI throughout the day and night. These factors strongly correlated with the cows’ movement towards or away from shaded areas, underscoring their relevance in managing thermal stress.
	
	However, some limitations have also been identified. The models struggled to accurately predict the exact number of cows under shade, suggesting that other variables not included in the study may also influence this behavior. Future research could explore these additional factors to enhance model performance.

	\section{Conclusions}\label{sec:conclusions}
	
	This study demonstrates the potential of using {\bf soft computing approaches for mathematical modeling} of noisy and highly variable biological behaviors.
	Using only climatic measurements and camera counts, both Random Forests and Neural Networks accurately predicted the number of dairy cows seeking shade during Mediterranean summer heat waves. The main conclusions are as follows:
	\begin{description}
		\item[Early warning capability:] the models anticipate shade-seeking peaks within one hour, with a median daily RMSE of  $13.84$ cows.
		\item[Interpretability:] a  $10$-tree Random Forest (depth $= 5$) achieves an average RMSE of $14.9$ while retaining a transparent rule structure, making it the recommended choice for on-farm deployment.
		\item[Minimal feature set:] three easily derived thermal features---current THI, accumulated daytime THI and mean night-time THI---are sufficient for a low-cost decision-support system that can trigger ventilation, sprinkling or shading strategies in real time.
	\end{description}
	
	These results show that soft computing models provide robust, affordable tools for precision-livestock management aimed at mitigating heat stress and safeguarding animal welfare.

	\vspace{1cm}
	
	\section*{Code Availability}
	All the algorithms presented in this paper are available in a GitHub repository. It can be accessed at the following link:
	
	\url{https://github.com/serjj99/CowShadeSeeking.git}.

	\section*{Acknowledgment}
	The research of J.M. Calabuig was funded by the Agencia Estatal de Investigaci\'on, grant number PID2022-138328OB-C21. 
	The research of Roger Arnau was funded by the Universitat Polit\`ecnica de Val\`encia, Programa de Ayudas de Investigaci\'on y Desarrollo (PAID-01-21).
	The research of Daniel A.M\'endez was supported by R+D+i project TED2021-130759B-C31, funded by MCIN/AEI/10.13039/501100011033/ and by the ``European Union NextGenerationEU/PRTR''.
	This work has received funding from the European Union's Horizon Europe research and innovation programme under the grant agreement No 01059609 (Re-Livestock project).
	The authors acknowledge the technicians of the Titaguas farm, Valencia (Spain), for their valuable help in the animal management during the research activities.

	\bibliographystyle{plain}
	\bibliography{biblio_shadow_Jose}

\end{document}